\documentclass{article}
\usepackage{caption}

\usepackage{booktabs}
\usepackage[table]{xcolor}
\usepackage{graphicx}
\usepackage{tabularx}
\usepackage{array}
\usepackage[table]{xcolor}
\usepackage{placeins}
\usepackage{float}
\usepackage{multirow}
\usepackage[preprint]{corl_2026}

\title{SAM3D-Guided Object-Centric Representation Alignment for Vision-Language-Action Models}

\author{
\vspace{0.8em}\\
\begin{tabular}{c}
{\normalsize
Zonghe Liu$^{*,1}$, ShanYuan Jie$^{*,2}$, Xiaoquan Sun$^{\ddagger,3}$, Chen Cao$^{\ddagger,1}$,}\\
{\normalsize
Zetian Xu$^{\ddagger,1}$, Liu Zongsheng$^{\ddagger,4}$, Jiayu Chen$^{\dagger,1,5}$}\\[0.25em]
{\small $^{1}$University of Hong Kong \quad
$^{2}$Shenzhen Institutes of Advanced Technology, Chinese Academy of Sciences}\\
{\small $^{3}$Huazhong University of Science and Technology \quad
$^{4}$Beijing University of Aeronautics and Astronautics}\\
{\small $^{5}$Infiforce}\\[0.1em]
{\small $^{*}$Equal Contribution \quad
$^{\ddagger}$Equal Second Contribution \quad
$^{\dagger}$Corresponding Author}
\end{tabular}
}
\usepackage{float}
\usepackage{graphicx}
\usepackage{amsmath}
\renewcommand{\thetable}{\arabic{table}}
\usepackage{chngcntr}
\counterwithout{table}{section}
\renewcommand{\thetable}{\arabic{table}}
\usepackage{enumitem}
\usepackage[most]{tcolorbox}

\usepackage{tcolorbox}

\begin{document}
\maketitle

\begin{figure}[H]
    \centering
    \includegraphics[width=1\linewidth,trim=0cm 0cm 0cm 0cm, clip]{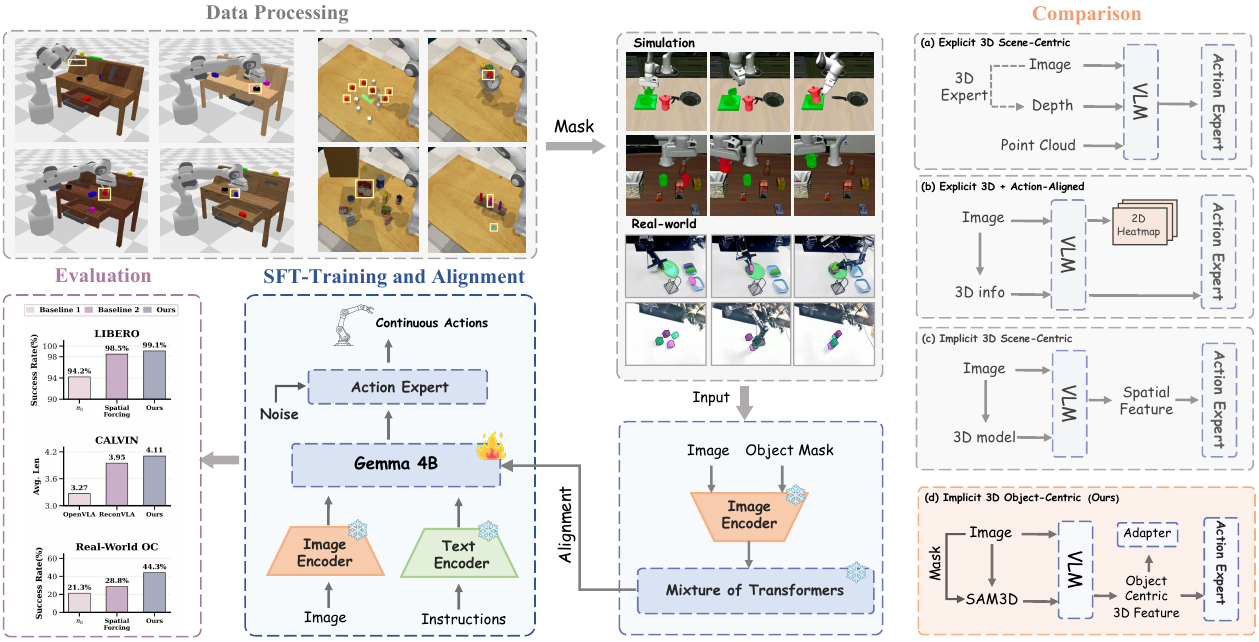}
        \caption{{\textbf{Overview of SAM3D-VLA}}. We propose an object-centric 3D alignment framework that uses SAM3D as a frozen teacher during training. High-level instructions are decomposed into subtasks, and task-relevant object masks are used for SAM3D feature extraction. The extracted 3D features are spatially resampled and dimensionally projected to align with intermediate VLA representations, while inference follows the original pipeline using RGB and language instructions.}

    \label{fig:framework}
\end{figure}

\begin{abstract}
Vision-Language-Action (VLA) models have shown strong potential for general robot manipulation, but most existing models rely on 2D visual-language backbones and lack fine-grained 3D understanding of target objects, especially under occlusion, pose variation, scale changes, and precise spatial interaction. We propose an object-centric 3D representation alignment framework built upon \(\pi_0\), using SAM3D as a frozen 3D teacher to provide target-object 3D priors during training. Specifically, we localize task-relevant objects with object recognition models, generate corresponding object masks, and use SAM3D to extract dense object-level 3D representations, which are aligned with intermediate visual features of \(\pi_0\). This enables the policy to internalize target-object 3D information while preserving the original RGB-language-to-action inference pipeline without requiring depth, point clouds, masks, SAM3D, or additional 3D modules at test time. Simulation experiments  show consistent improvements, achieving \(99.1\%\) on LIBERO and \(4.11\) average length on CALVIN. Real-world experiments further demonstrate that our method is particularly effective in long-horizon manipulation scenarios where the robot must focus on different target objects across multiple subtasks.
\end{abstract}



\section{Introduction}

Vision-Language-Action (VLA) models have become an important direction for general robot manipulation, as they adapt pretrained vision-language models to predict robot actions~\cite{rt2, openvla, pi_0}. With large-scale visual-language pretraining, these models can understand language instructions and handle a wide range of manipulation tasks. However, most existing VLAs still rely mainly on 2D RGB images and learn from action prediction alone. As a result, they may not capture the 3D properties of the target object, such as its shape, pose, scale, and spatial layout. This becomes a clear limitation in tasks with occlusion, pose changes, clutter, or precise object placement.

Recent works have attempted to enhance VLA models with spatial or 3D information. Some methods introduce explicit 3D observations, such as depth maps, point clouds, or RGB-D inputs, into the policy~\cite{pointvla, bridgevla, geovla, 3d}. Others inject estimated 3D positions, spatial action grids, or representation-level spatial supervision into VLA models~\cite{spatialvla, spatial}. These methods demonstrate the importance of 3D structure for manipulation, but they often require additional 3D inputs, modify the original input-output interface, or focus mainly on global scene-level spatial representations. Meanwhile, visual grounding methods improve target-region attention through cropped inputs, bounding-box prediction, or gaze-region reconstruction~\cite{reconvla, roboground, ECoT, graspvla}, but their supervision remains primarily in the 2D image space and does not explicitly provide target-object 3D shape or layout priors.

This raises a key question: can we inject object-centric 3D knowledge into an RGB-based VLA policy while preserving its original inference pipeline? To this end, we propose a SAM3D-guided object-centric 3D representation alignment framework built upon \(\pi_0\)~\cite{pi_0}. During training, we automatically ground the task-relevant object with open-vocabulary detection~\cite{yolov12, grounding} and SAM2~\cite{sam2} segmentation, and use a frozen SAM3D~\cite{sam3d} teacher to extract dense object-centric 3D features from the masked target. These teacher features are spatially resampled to match the visual token grid of \(\pi_0\), and the intermediate VLA features are projected into the SAM3D feature space for masked normalized representation alignment. For long-horizon tasks, we further decompose high-level instructions into subtasks and associate each subtask with its corresponding target object, providing stage-specific 3D supervision. Importantly, all teacher-side modules are used only during training. 

During inference, our policy follows the original \(\pi_0\) RGB-language-to-action pipeline and requires no depth maps, point clouds, object masks, SAM3D, or additional 3D modules. Experiments on LIBERO and CALVIN show that our method consistently improves over strong VLA baselines, achieving \textbf{99.1\%} on LIBERO and average length of \textbf{4.11} on CALVIN. We further verify the mechanism with a frozen-representation probing experiment, showing that SAM3D object-centric 3D priors can be more accurately predicted from the learned VLA features after training.

Our contributions are summarized as follows:
\begin{itemize}[
    labelindent=1em,
    leftmargin=*,
    itemsep=0pt,
    topsep=0pt,
    parsep=0pt,
    partopsep=0pt
]
    \item We identify Object-Centric 3D understanding as a key bottleneck of RGB-based VLA policies, complementing prior works that mainly focus on global spatial awareness, action-space spatialization, or 2D visual grounding.
    \item We propose \textbf{SAM3D-VLA} that distills shape and layout priors of target object into the intermediate visual features of \(\pi_0\)~\cite{pi_0}, while preserving the original inference pipeline without depth maps, point clouds, masks, or additional 3D modules at test time.
    \item We demonstrate consistent improvements on simulation and real-world tasks, and further verify the mechanism through a frozen-representation probing experiment, showing that object-centric 3D priors can be more accurately recovered from the learned VLA features after training.
\end{itemize}

\section{Related Works}

\textbf{Vision-Language-Action Models}.  Vision-Language-Action (VLA) models have become a promising paradigm for general robot manipulation by adapting pretrained vision-language backbones to robot control~\cite{rt2,openvla,pi_0}. These models can follow language instructions and perform diverse manipulation tasks, but most of them rely mainly on 2D RGB observations, limiting their ability to capture object-level 3D structure. Visual grounding methods improve target attention through cropped regions, bounding boxes, or gaze reconstruction~\cite{reconvla,roboground,vip,ECoT,graspvla}. However, their supervision remains mostly in the 2D image space and does not explicitly provide target-object 3D shape or layout priors.

\textbf{Spatial and 3D-aware VLA Models}. Another line of work introduces spatial or 3D information into VLA models to improve manipulation performance. Some methods incorporate explicit 3D observations such as depth maps, point clouds, or RGB-D inputs~\cite{3d,pointvla,vidbot,bridgevla}. For example, BridgeVLA projects point clouds into orthographic 2D views and predicts heatmaps for action localization, but it still relies on 3D observations and changes the policy input-output formulation. Other methods inject spatial information through architectural designs, action representations, or representation-level supervision. SpatialVLA~\cite{spatialvla} uses Ego3D Position Encoding and Adaptive Action Grids, while Spatial Forcing~\cite{spatial} aligns intermediate VLA features with representations from pretrained 3D foundation models. These works show the value of spatial structure for manipulation, but often focus on explicit 3D inputs, action-space redesign, or scene-level spatial supervision. In contrast, our method uses subtask-aware object-centric 3D priors only during training, while preserving the original RGB-language-to-action inference pipeline of \(\pi_0\).

\section{Method}

\subsection{SAM3D-guided VLA Training Framework}

\paragraph{Vision-Language-Action Models.}
SAM3D-VLA is built upon the \(\pi_0\)~\cite{pi_0} architecture, which combines a pretrained vision-language backbone with a continuous action expert. Following \(\pi_0\), the VLM backbone is composed of a SigLIP~\cite{siglip} vision encoder and a Gemma~\cite{gemma} language model. Formally, at robot timestep \(t\), we model the conditional action distribution $p_\theta(\mathbf{A}_t \mid \mathbf{o}_t)$, where \(\mathbf{A}_t=[\mathbf{a}_t,\mathbf{a}_{t+1},\ldots,\mathbf{a}_{t+H-1}]\) denotes an action chunk of horizon \(H\), and \(\mathbf{o}_t\) is the robot observation. The observation consists of multi-view RGB images, a language instruction, and the robot proprioceptive state: $\mathbf{o}_t = [\mathbf{I}_t^1,\ldots,\mathbf{I}_t^n,\ell,\mathbf{q}_t],$
where \(\mathbf{I}_t^i\) is the \(i\)-th camera image, \(\ell\) is the language command, and \(\mathbf{q}_t\) denotes the robot state such as joint angles or gripper state.

The RGB images are encoded by the SigLIP vision encoder into visual tokens, while the language command is embedded into text tokens in the Gemma token space. The robot state is projected into the same embedding dimension through a linear projection layer. These tokens are then processed by the Gemma-based multimodal Transformer:
\begin{equation}
    \mathbf{H}_t = f_{\mathrm{VLM}}(\mathbf{I}_t^1,\ldots,\mathbf{I}_t^n,\ell,\mathbf{q}_t).
\end{equation}
where \(\mathbf{H}_t=\{\mathbf{h}_t^{(1)},\ldots,\mathbf{h}_t^{(L)}\}\) denotes hidden representations from different Transformer layers. In the standard \(\pi_0\) pipeline, these features condition the action expert for continuous action generation.

For action prediction, each action in the chunk is represented as a continuous action token and processed by the action expert. We train the action expert using a conditional flow matching objective. Given a ground-truth action chunk \(\mathbf{A}_t\), we sample Gaussian noise \(\boldsymbol{\epsilon}\sim\mathcal{N}(\mathbf{0},\mathbf{I})\) and a flow matching timestep \(\tau\in[0,1]\). The noisy action chunk is constructed as $\mathbf{A}_t^\tau = \tau \mathbf{A}_t + (1-\tau)\boldsymbol{\epsilon}$,  the action expert predicts a velocity field conditioned on the VLM features: $\hat{\mathbf{v}}_\theta
=
\mathbf{v}_\theta(\mathbf{A}_t^\tau,\tau,\mathbf{o}_t)$, and is optimized to match the target denoising direction:
\begin{equation}
    \mathcal{L}_{\mathrm{action}}
=
\mathbf{E}_{\tau,\boldsymbol{\epsilon}}
\left[
\left\|
\mathbf{v}_\theta(\mathbf{A}_t^\tau,\tau,\mathbf{o}_t)
-
(\mathbf{A}_t-\boldsymbol{\epsilon})
\right\|_2^2
\right].
\end{equation}
During inference, actions are generated by starting from Gaussian noise and integrating the learned velocity field from \(\tau=0\) to \(\tau=1\), producing the final action chunk for robot execution.

In our framework, we preserve the original \(\pi_0\) formulation, where the policy maps RGB observations and language instructions to continuous actions. The key difference is that, during training, we additionally extract the intermediate visual features \(\mathbf{h}_t^{(m)}\) from a selected Gemma layer and project them into the SAM3D representation space. These projected VLA features are supervised by object-centric 3D shape and layout priors from SAM3D. This design injects 3D object knowledge into the VLA representation while keeping the original \(\pi_0\) inference pipeline unchanged.

\begin{figure}[t]
    \centering
    \includegraphics[width=1\linewidth,trim=0cm 0cm 0cm 0cm, clip]{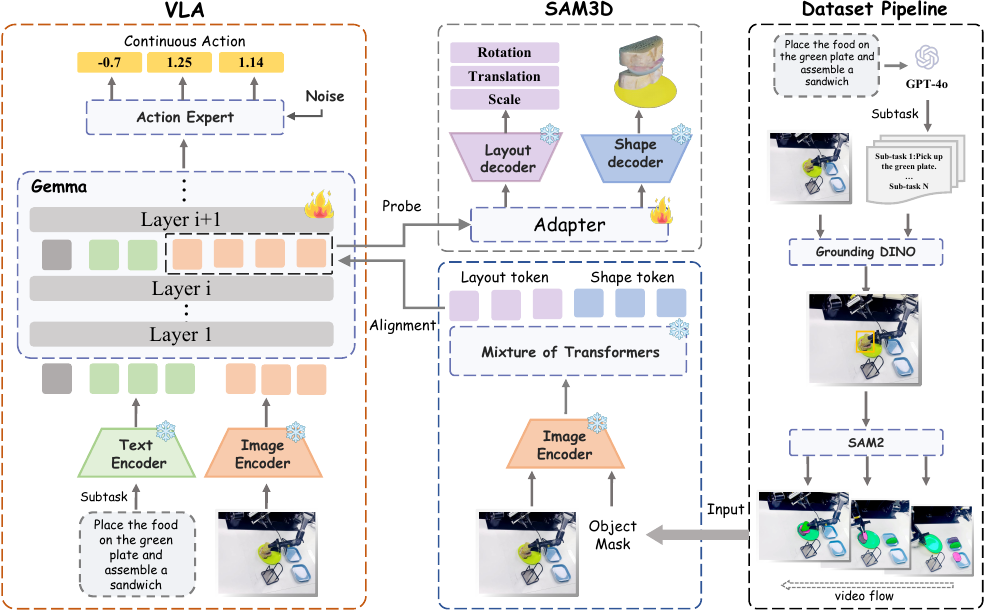}
    \caption{{\textbf{The framework of SAM3D-VLA}.} Task-relevant object masks are generated from subtask instructions and fed to a frozen SAM3D teacher to provide object-centric 3D supervision for intermediate VLA features. We further use a frozen-representation probing adapter to verify whether SAM3D priors are encoded in the learned VLA representations. }
    \label{fig:framework}
\end{figure}

\paragraph{SAM3D-guided Object-centric Feature Alignment.}
We use SAM3D as a frozen 3D teacher to provide object-centric geometric supervision for the intermediate visual features of the \(\pi_0\) backbone. Since SAM3D operates on single images, we flatten a batch of multi-view observations 
\(\mathbf{I}_t \in \mathbf{R}^{B \times V \times 3 \times H \times W}\)
into 
\(\tilde{\mathbf{I}}_t \in \mathbf{R}^{(BV) \times 3 \times H \times W}\),
where \(B\) is the batch size, \(V\) is the number of camera views, and \(H,W\) denote the image resolution. We flatten the corresponding subtask-specific object masks in the same way. The image-mask pairs are fed into the frozen SAM3D teacher to extract dense object-centric 3D features: $
\mathbf{T}_t = f_{\mathrm{SAM3D}}(\tilde{\mathbf{I}}_t, \tilde{\mathbf{M}}_t)$, $\mathbf{T}_t \in \mathbf{R}^{(BV) \times L_T \times D_T}$, where \(L_T\) is the number of SAM3D teacher tokens and \(D_T\) is the teacher feature dimension. These features are extracted from the last transformer block of SAM3D and contain object-level geometric priors such as shape, surface structure, and spatial layout.

Because SAM3D and \(\pi_0\) have different token resolutions, we reshape the teacher sequence into a 2D feature grid, $\mathbf{T}_t \rightarrow \mathbf{T}^{2D}_t \in \mathbf{R}^{(BV) \times D_T \times H_T \times W_T}$, remove the global token if it exists, and resize the grid to the student visual-token resolution by bilinear interpolation: $\bar{\mathbf{T}}^{2D}_t
=\mathrm{Interp}(\mathbf{T}^{2D}_t, H_S, W_S)$, $\quad H_S W_S = L_S$, where \(L_S\) is the number of student visual tokens per camera view. The resized features are flattened and reassembled across views as $\bar{\mathbf{T}}_t \in \mathbf{R}^{B \times (V L_S) \times D_T}$, making the teacher tokens spatially aligned with the corresponding \(\pi_0\) visual tokens. On the student side, we extract intermediate visual features from a selected Gemma layer: $\mathbf{S}_t = \mathbf{h}^{(m)}_t \in \mathbf{R}^{B \times (V L_S) \times D_S}$, where \(D_S\) is the student feature dimension. Since the feature dimensions differ, we project the student features into the SAM3D feature space: $\hat{\mathbf{T}}_t = P_\phi(\mathbf{S}_t)$, $\hat{\mathbf{T}}_t \in \mathbf{R}^{B \times (V L_S) \times D_T}$.

To focus supervision on task-relevant regions, we construct a token-level mask 
\(\mathbf{m}_t \in \{0,1\}^{B \times (V L_S)}\)
from the subtask-specific object masks and apply alignment only to target-object tokens. We further denote by \(\mathbf{M}_t \in \{0,1\}^{B \times (V L_S) \times D_T}\) the mask expanded along the feature dimension. Both projected student features and SAM3D teacher features are \(\ell_2\)-normalized, and the object-centric alignment loss is defined as a masked normalized MSE:
\begin{equation}
\mathcal{L}_{\mathrm{align}}
=
\mathrm{MSE}
\left(
\mathrm{Norm}(\hat{\mathbf{T}}_t)[\mathbf{M}_t],
\mathrm{Norm}(\bar{\mathbf{T}}_t)[\mathbf{M}_t]
\right).
\end{equation}
This normalized objective encourages directional feature alignment and is robust to scale differences between SAM3D and \(\pi_0\). The final training objective combines the standard \(\pi_0\) flow-matching action loss with the SAM3D-guided alignment loss:
\begin{equation}
\mathcal{L}
=
\mathcal{L}_{\mathrm{action}}
+
\alpha \mathcal{L}_{\mathrm{align}} .
\label{eq:total_loss}
\end{equation}
where \(\alpha\) controls the strength of the 3D feature supervision. This objective encourages the VLA backbone to encode object-centric 3D priors while preserving its original action prediction ability.

\paragraph{Frozen-representation Probing Adapter.}

To verify whether SAM3D-guided alignment makes object-centric 3D information more accessible in the learned VLA representation, we conduct a frozen-representation probing experiment. After policy training, we freeze the entire VLA backbone and train only a lightweight two-layer MLP probe \(P_\omega\) on top of the same intermediate visual features \(\mathbf{S}_t\) used for alignment. The probe predicts the spatially resampled SAM3D target \(\bar{\mathbf{T}}_t\): $\tilde{\mathbf{T}}_t = P_\omega(\mathbf{S}_t)$, $\tilde{\mathbf{T}}_t \in \mathbf{R}^{B \times (V L_S) \times D_T}$. Using the same expanded object mask \(\mathbf{M}_t\), the probe is trained with
\begin{equation}
    \mathcal{L}_{\mathrm{probe}}
=
\mathrm{MSE}
\left(
\mathrm{Norm}(\tilde{\mathbf{T}}_t)[\mathbf{M}_t],
\mathrm{Norm}(\bar{\mathbf{T}}_t)[\mathbf{M}_t]
\right).
\end{equation}
Only \(P_\omega\) is optimized, while the VLA model remains frozen. Therefore, better probing performance indicates that SAM3D-style object-centric 3D priors are more recoverable from the learned VLA features. We compare the original \(\pi_0\) baseline with our SAM3D-aligned model to quantify this effect.

\subsection{Subtask-aware Data Processing}


For each training frame and camera view, we localize the subtask-relevant object using an object detection model and then use SAM2 to generate its binary mask. The resulting image-mask pairs are fed into the frozen SAM3D teacher to extract object-centric 3D features. Since SAM3D processes single images, multi-view observations are flattened along the camera dimension and each view is processed independently. These SAM3D features are used as training-time targets for representation alignment. During inference, this entire data processing pipeline is removed, and the policy follows the original \(\pi_0\) pipeline using only RGB observations and language instructions.



	
\begin{table*}[t]
\centering
\caption{Comparisons with state-of-the-art methods on LIBERO benchmark. \textbf{Bold} denotes the best performance among all methods.}
\label{tab:success_rate}
\resizebox{\textwidth}{!}{
\begin{tabular}{lccccc}
\toprule
Method &
Spatial SR (\%) &
Object SR (\%) &
Goal SR (\%) &
Long SR (\%) &
Average SR (\%) \\
\midrule

\rowcolor{gray!15}
\multicolumn{6}{c}{2D VLA} \\
\midrule
Diffusion Policy~\cite{diffusion}
& 78.3 & 92.5 & 68.3 & 50.5 & 72.4 \\
Octo~\cite{octo}
& 78.9 & 85.7 & 84.6 & 51.1 & 75.1 \\
OpenVLA~\cite{openvla}
& 84.7 & 88.4 & 79.2 & 53.7 & 76.5 \\
Dita~\cite{dita}
& 84.2 & 96.3 & 85.4 & 63.8 & 82.4 \\
CoT-VLA~\cite{cot}
& 87.5 & 91.6 & 87.6 & 69.0 & 83.9 \\
$\pi_0$~\cite{pi_0}
& 96.8 & 98.8 & 95.8 & 85.2 & 94.2 \\
UniVLA~\cite{univla}
& 96.5 & 96.8 & 95.6 & 92.0 & 95.2 \\
OpenVLA-OFT~\cite{OpenVLA-OFT}
& 97.6 & 98.4 & 97.9 & 94.5 & 97.1 \\

\midrule
\rowcolor{gray!15}
\multicolumn{6}{c}{Explicit 3D VLA} \\
\midrule
SpatialVLA~\cite{spatialvla}
& 88.2 & 89.9 & 78.6 & 55.5 & 78.1 \\
GeoVLA~\cite{geovla}
& 98.4 & 99.0 & 96.6 & 96.6 & 97.7 \\
3D-CAVLA~\cite{3d}
& 98.2 & \textbf{99.8} & 98.2 & 96.1 & 98.1 \\

\midrule
\rowcolor{gray!15}
\multicolumn{6}{c}{Implicit 3D VLA} \\
\midrule
Spatial Forcing~\cite{spatial}
&\textbf{99.4} & 99.6 & 98.8 & 96.0 & 98.5\\
\rowcolor{purple!10}
SAM3D-VLA (Ours)
&99.2 & 99.7 & \textbf{99.1} & \textbf{98.4} & \textbf{99.1} \\
\bottomrule
\end{tabular}
}
\end{table*}

\begin{table*}[t]
\centering
\caption{Comparisons with state-of-the-art methods on CALVIN benchmark. \textbf{Bold} denotes the best performance among all methods.}
\label{tab:long_horizon_results}
\footnotesize
\setlength{\tabcolsep}{5pt}
\renewcommand{\arraystretch}{0.9}

\begin{tabularx}{\textwidth}{l c *{6}{>{\centering\arraybackslash}X}}
\toprule
Method & Splits
& \multicolumn{5}{c}{Success Rate (\%)} 
& Avg. Len \\
\cmidrule(lr){3-7}
& & 1/5 & 2/5 & 3/5 & 4/5 & 5/5 & \\
\midrule

\rowcolor{gray!15}
\multicolumn{8}{c}{Generative Methods} \\
\midrule
UniPi~\cite{unipi}
& ABC$\rightarrow$D & 56.0 & 16.0 & 8.0 & 8.0 & 4.0 & 0.92 \\
SuSIE~\cite{SuSIE}
& ABC$\rightarrow$D & 87.0 & 69.0 & 49.0 & 38.0 & 26.0 & 2.69 \\
GR-1~\cite{GR-1}
& ABC$\rightarrow$D & 85.4 & 71.2 & 59.6 & 49.7 & 40.1 & 3.06 \\
VidMan~\cite{vidman}
& ABC$\rightarrow$D & 91.5 & 76.4 & 68.2 & 59.2 & 46.7 & 3.42 \\
CLOVER~\cite{Clover}
& ABC$\rightarrow$D & 96.0 & 83.5 & 70.8 & 57.5 & 45.4 & 3.53 \\

\midrule
\rowcolor{gray!15}
\multicolumn{8}{c}{Large VLA Models} \\
\midrule
VLAS~\cite{vlas}
& ABC$\rightarrow$D & 87.2 & 64.2 & 40.9 & 28.1 & 19.6 & 2.40 \\
RoboFlamingo~\cite{RoboFlamingo}
& ABC$\rightarrow$D & 82.4 & 61.9 & 46.6 & 33.1 & 23.5 & 2.47 \\
OpenVLA~\cite{openvla}
& ABC$\rightarrow$D & 91.3 & 77.8 & 62.0 & 52.1 & 43.5 & 3.27 \\
UniVLA~\cite{univla}
& ABC$\rightarrow$D & 95.5 & 85.8 & 75.4 & 66.9 & 56.5 & 3.80 \\

\midrule
\rowcolor{gray!15}
\multicolumn{8}{c}{Object-centric Methods} \\
\midrule
ReconVLA~\cite{reconvla}
& ABC$\rightarrow$D & 95.6 & 87.6 & 76.9 & 69.3 & 64.1 & 3.95 \\

\rowcolor{purple!10}
SAM3D-VLA (Ours)
& ABC$\rightarrow$D & \textbf{96.2} & \textbf{89.1} & \textbf{80.5} & \textbf{73.6} & \textbf{71.6} & \textbf{4.11} \\

\bottomrule
\end{tabularx}
\end{table*}

\section{Experiments}
\paragraph{Simulation Experiments on LIBERO.}
We evaluate our method on LIBERO benchmarks~\cite{libero}. LIBERO contains LIBERO-Spatial, LIBERO-Object, LIBERO-Goal, and LIBERO-Long. Each task suite contains 500 expert demonstrations across 10 tasks, designed to probe generalization to different spatial layouts, objects, goals, and long-horizon behaviors. SAM3D-VLA is compared with representative 2D VLA models~\cite{diffusion, tracevla, octo, openvla, dita, cot, pi_0, pi_0fast}, recent 3D or spatially enhanced VLA methods~\cite{spatialvla, geovla, 3d} and Spatial Forcing~\cite{spatial}. All methods are evaluated using the standard success-rate metrics of each benchmark.

\paragraph{Experiments results.}
As shown in Table~\ref{tab:success_rate}, SAM3D-VLA achieves the best average success rate on LIBERO, reaching \textbf{99.1\%}. Compared with strong 2D VLA baselines such as \(\pi_0\), UniVLA, and OpenVLA-OFT, our method obtains clear gains across object, goal, and long-horizon task categories. The improvement is especially evident on LIBERO-Long, where our method reaches \textbf{98.4\%}. Since each LIBERO-Long task contains two sequential subtasks, the task-relevant object may change across stages. Our subtask-aware processing associates each subtask with its corresponding object mask, providing stage-specific SAM3D supervision and helping the policy focus on different target objects during different manipulation phases.

\paragraph{Simulation Experiments on CALVIN.}
CALVIN benchmark~\cite{calvin} consists of 34 tasks and 4 different environments (A, B, C and D). The CALVIN long-horizon challenge is a sequential task comprising five subtasks. We report the success rates for each subtask and the average completed length across all five tasks. The method is evaluated over 500 rollouts to ensure a fair comparison. The metrics of CALVIN are the success rates of each subtask and the average length of all sequential 5 subtasks.

\paragraph{Experiments results.}

On CALVIN, as shown in Table~\ref{tab:long_horizon_results}, SAM3D-VLA achieves the best performance with an average length of \textbf{4.11} and a 5-step success rate of \textbf{71.6\%}. Compared with ReconVLA, UniVLA, OpenVLA, and other baselines, our method consistently improves success rates across task lengths, showing stronger long-horizon consistency. These gains indicate that SAM3D-guided object-centric alignment helps the policy encode target-object 3D priors for sequential manipulation, while preserving the original \(\pi_0\) RGB-language-to-action inference pipeline without additional test-time inputs or modules.

\begin{figure}[t]
    \centering
    \includegraphics[width=1\linewidth,trim=0cm 0cm 0cm 0cm, clip]{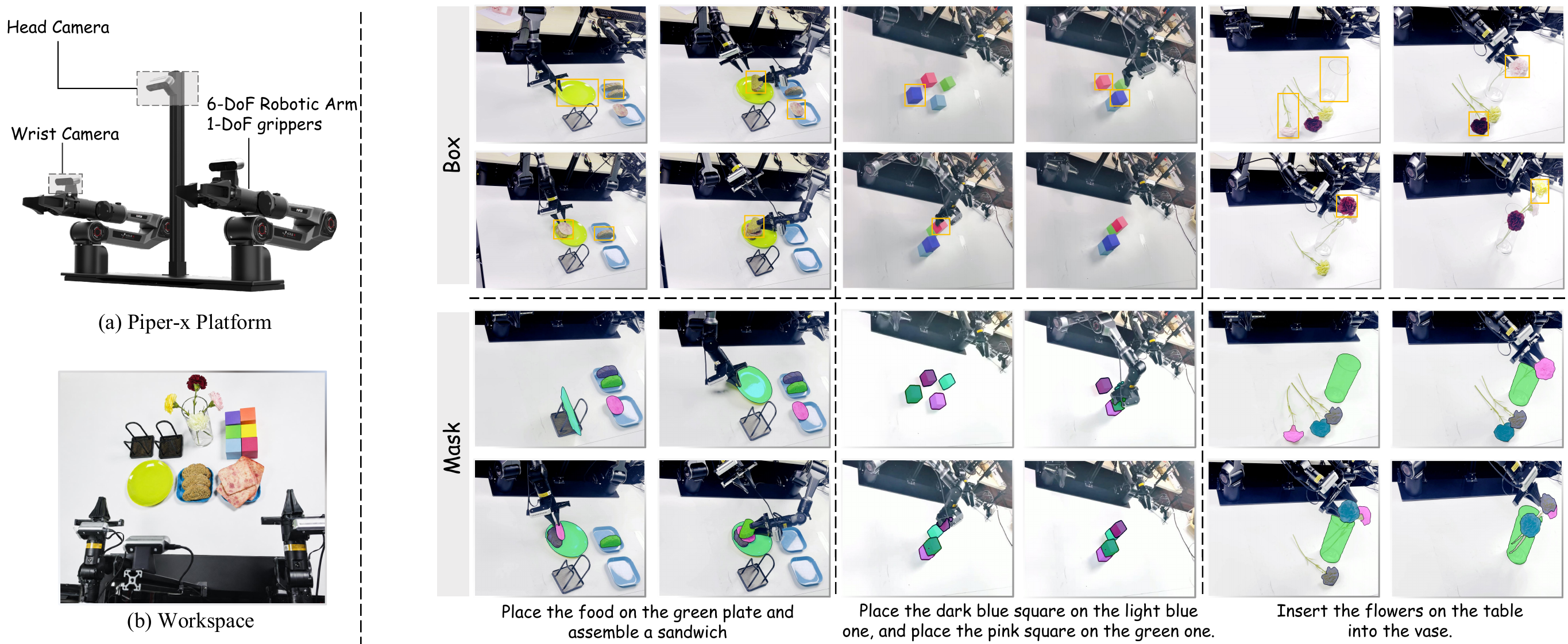}
    \caption{{\textbf{Real-world experimental setup and Mask processing}.}  (a) Piper-x Platform. (b) Workspace. Dataset Processing: The target object is grounded and masked by open-source models. }
    \label{fig:Real}
\end{figure}

\paragraph{Real-world setup.}
We further evaluate our method on the Piper-X robotic arm, a dual-arm mobile platform equipped with three omnidirectional wheels, two 6-DoF arms, two wrist cameras, and a head camera, as shown in Fig.~\ref{fig:Real}. During deployment, the policy follows the original \(\pi_0\) inference pipeline, taking only RGB observations and language instructions as input to predict continuous action chunks. We conduct experiments on three real-world manipulation scenarios: cooking, flower arrangement, and block stacking, covering object selection, grasping, insertion and placement. All methods are evaluated under the same robot setup and protocol, and training is conducted on 8 NVIDIA H100 GPUs.

\begin{figure}[t]
    \centering
    \includegraphics[width=0.9\linewidth,trim=0cm 0cm 0cm 0cm, clip]{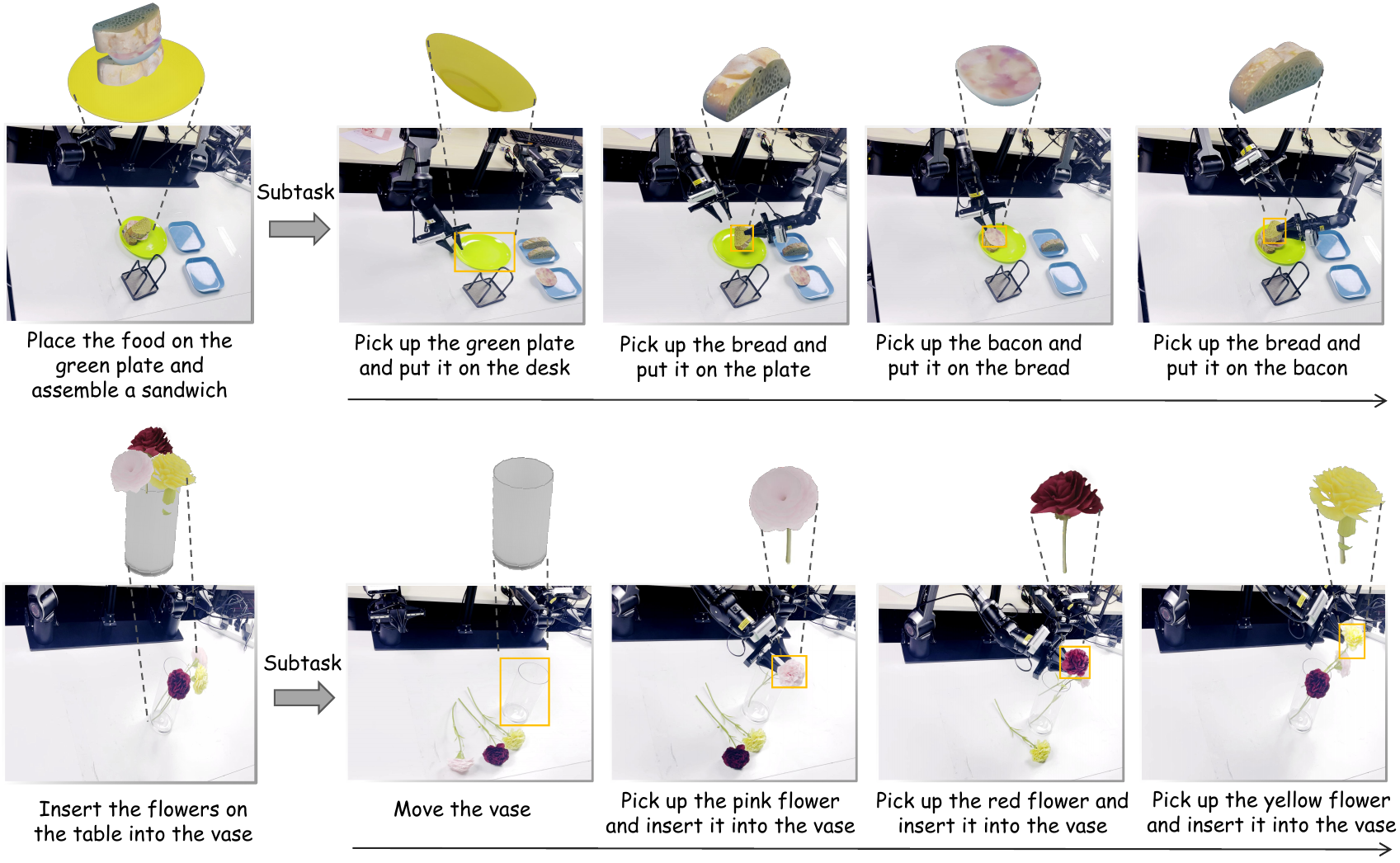}
    \caption{{\textbf{Subtask-aware object-centric data processing}.}  High-level long-horizon instructions are decomposed into subtasks, and each subtask is associated with its corresponding target object mask. During training, the model receives stage-specific SAM3D supervision, encouraging it to focus on different task-relevant objects across manipulation steps.}
    \label{fig:subtask}
\end{figure}

\noindent
\begin{minipage}[t]{0.56\linewidth}
\vspace{0pt}
\paragraph{Real-world data processing.}
For real-world training data, we use the same subtask-aware object-centric pipeline as in simulation. As showen in Fig.~\ref{fig:subtask}, each high-level instruction is decomposed into subtasks by an LLM, and the corresponding target objects are grounded by Grounding DINO~\cite{grounding} and YOLO~\cite{yolov12}, followed by SAM2~\cite{sam2} segmentation. The resulting image-mask pairs are fed into the frozen SAM3D~\cite{sam3d} teacher to extract object-centric 3D features for training-time alignment. During deployment, no subtask decomposition, detection, segmentation, or SAM3D module is required.
\end{minipage}%
\hfill%
\begin{minipage}[t]{0.40\linewidth}
\vspace{0pt}
\centering
\refstepcounter{table}
\label{tab:Real_results}
\parbox{\linewidth}{
\centering
\small
Table~\thetable: Real-world results.
}
\vspace{-10pt}

\setlength{\tabcolsep}{3pt}
\renewcommand{\arraystretch}{0.9}
\resizebox{\linewidth}{!}{
\begin{tabular}{lcccc}
\toprule
\multirow{2}{*}{\textbf{Task}} 
& \multicolumn{2}{c}{\(\boldsymbol{\pi_0}\)~\cite{pi_0}} 
& \multicolumn{2}{c}{\textbf{SAM3D-VLA}} \\
\cmidrule(lr){2-3} \cmidrule(lr){4-5}
& ST & OC & ST & OC \\
\midrule
\multicolumn{5}{l}{\textbf{Basic Tasks}} \\
Stack bowls & 90\% & 55\% & \textbf{92\%} & \textbf{81\%} \\
Hang cup    & 75\% & 15\% & \textbf{81\%} & \textbf{36\%} \\
Open drawer & \textbf{68\%} & 30\% & 62\% & \textbf{40\%} \\
\midrule
\multicolumn{5}{l}{\textbf{Long Tasks}} \\
Cook   & 15\% & 5\%  & \textbf{40\%} & \textbf{26\%} \\
Flower & 30\% & 13\% & \textbf{62\%} & \textbf{45\%} \\
Stack  & 23\% & 10\% & \textbf{54\%} & \textbf{38\%} \\
\midrule
\textbf{Average} & 50.2\% & 21.3\% & \textbf{65.2\%} & \textbf{44.3\%} \\
\bottomrule
\end{tabular}
}
\end{minipage}

\paragraph{Real-world results.}
Table~\ref{tab:Real_results} reports the real-world success rates on the AgileX PiperX robot. We evaluate each task under two settings: the standard setting (ST), where objects are placed in canonical initial states, and the occlusion setting (OC), where we introduce additional disturbances such as changed object positions, irrelevant distractor objects, and partial occlusions. SAM3D-VLA improves the average success rate from \(50.2\%\) to \textbf{65.2\%} under ST and from \(21.3\%\) to \textbf{44.3\%} under OC, with clear gains on long-horizon tasks: cooking, inserting flower and stacking. These results suggest that subtask-aware SAM3D supervision helps the policy focus on the correct target object at each stage and encode object-level shape and layout priors, improving real-world robustness while preserving the original RGB-language-to-action inference pipeline without masks, SAM3D, depth maps, or point clouds during deployment.


\section{Conclusion}
\label{sec:conclusion}

In this paper, we propose SAM3D-VLA, an object-centric 3D representation alignment framework for vision-language-action models. By using a frozen SAM3D teacher during training, our method distills target-object 3D priors into intermediate VLA features while preserving the original RGB-language-to-action inference pipeline. For long-horizon tasks, we further decompose high-level instructions into subtasks and provide stage-specific object-centric supervision. Experiments on LIBERO, CALVIN, and real-world Piper-X tasks show that our method consistently improves manipulation performance, especially under long-horizon and object-centric generalization settings. These results suggest that training-time object-level 3D priors are an effective way to enhance VLA policies without requiring depth, point clouds, masks, or 3D modules during deployment.

\section{Limitations}

Although SAM3D-VLA improves VLA policies with object-centric 3D priors, it still has limitations. The training pipeline depends on automatically generated subtask annotations and object masks, so errors in decomposition, grounding, or segmentation may introduce noisy supervision. SAM3D also operates on single images, which may limit its reliability under severe occlusion, transparent objects, or poor viewpoints. In addition, our real-world evaluation is limited to tabletop tasks on one robot platform. While this work builds upon \(\pi_0\), the proposed alignment framework is not restricted to it. Future work will improve subtask-object association, explore multi-view or temporal 3D teachers, and evaluate the method on other mainstream VLA backbones and broader robot settings.

\clearpage


\bibliography{corl_2026_template_cameraready/main}  

\end{document}